\documentclass[10pt,twocolumn,letterpaper]{article}

\usepackage{wacv}
\usepackage{times}
\usepackage{epsfig}
\usepackage{graphicx}
\usepackage{amsmath}
\usepackage{amssymb}
\usepackage{booktabs}
\usepackage{textcomp}
\usepackage[table,xcdraw]{xcolor}
\usepackage{caption}
\usepackage{subcaption}
\usepackage{multirow}
\usepackage{hhline}
\usepackage{algorithm}
\usepackage{algorithmic}
\usepackage{nicefrac}

\newcommand\convname{HyveConv}

\newcommand\blfootnote[1]{%
  \begingroup
  \renewcommand\thefootnote{}\footnote{#1}%
  \addtocounter{footnote}{-1}%
  \endgroup
}

\def\wacvPaperID{1445} 

\wacvalgorithmstrack   

\wacvfinalcopy 

\ifwacvfinal
\usepackage[breaklinks=true,bookmarks=false]{hyperref}
\else
\usepackage[pagebackref=true,breaklinks=true,colorlinks,bookmarks=false]{hyperref}
\fi

\begin{document}

\title{Wavelength-aware 2D Convolutions for Hyperspectral Imaging}

\author{Leon Amadeus Varga, Martin Messmer, Nuri Benbarka, Andreas Zell\\
Cognitive Systems Group\\
University of Tuebingen\\
Tuebingen, Germany\\
{\tt\small leon.varga@uni-tuebingen.de, martin.messmer@uni-tuebingen.de,}\\ {\tt\small nuri.benbarka@uni-tuebingen.de, andreas.zell@uni-tuebingen.de}
}

\maketitle
\thispagestyle{empty}

\begin{abstract}
    Deep Learning could drastically boost the classification accuracy for Hyperspectral Imaging (HSI). Still, the training on the mostly small hyperspectral data sets is not trivial. Two key challenges are the large channel dimension of the recordings and the incompatibility between cameras of different manufacturers.
    
    By introducing a suitable model bias and continuously defining the channel dimension, we propose a 2D convolution optimized for these challenges of Hyperspectral Imaging.
    We evaluate the method based on two different hyperspectral applications (inline inspection and remote sensing). Besides the shown superiority of the model, the modification adds additional explanatory power.
    
    In addition, the model learns the necessary camera filters in a data-driven manner. Based on these camera filters, an optimal camera can be designed.
\end{abstract}
\blfootnote{A PyTorch implementation of the model and supplementary material is available at:\\ \url{https://github.com/cogsys-tuebingen/hyve\_conv}}

\section{Introduction}
Image classification is one of the main tasks in computer vision \cite{deng2009imagenet}. Recent approaches could outperform humans, and this problem seems nearly solved for common object scenarios. These research findings were utilized for more complex tasks like image segmentation or object detection \cite{DBLP:conf/cvpr/HeZRS16}.

Image classification focuses primarily on color images. Color images often consist of three channels (red, green, and blue) and cover only the visible spectrum of light. Therefore, they can mimic human perception.

\begin{figure}
    \centering
    \includegraphics[width=0.9\columnwidth]{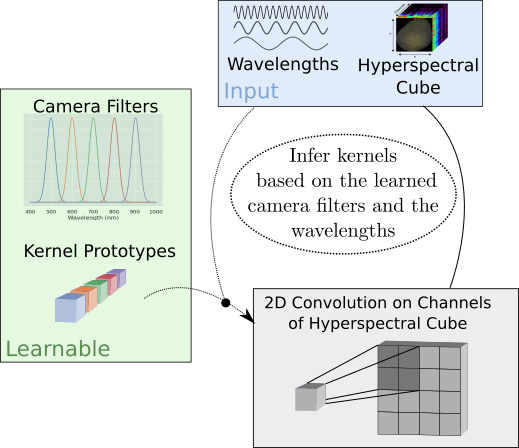}
    \caption{Hyperspectral Visual Embedding Convolution (HyveConv) at a glance. Further details in section \ref{sec:method}.}
    \label{fig:procedure}
\end{figure}

Hyperspectral recordings approximate the spectrum for each pixel of the image. Therefore, the number of channels is increased (to around 200), and the range of recorded wavelengths is extended. The additional wavelengths can carry information, which is helpful for complex classification tasks. These systems can perform tasks that aren't possible with pure human perception, allowing superhuman performance in several applications.

The evolution of hyperspectral camera systems decreased their price and simplified their usage. This enabled their application in many new areas, e.g. food processing and agricultur \cite{park2015hyperspectral}\cite{yang2011}, medical applications \cite{10.1117/1.JBO.19.1.010901}, inline sorting \cite{TATZER200599}, besides their original use in remote sensing \cite{PURR1947}.

The application of hyperspectral cameras often requires complex data acquisition (e.g., line-scan operation mode) and labeling procedures. This is leading to small data sets. The small data sets and the complicated features, which are often necessary for the tasks, support overfitting. Besides these characteristics,  the larger channel dimension of hyperspectral recordings requires special attention. To tackle this issue, using dimension reduction techniques, such as PCA \cite{doi:10.1080/14786440109462720} or Factor Analysis \cite{thurstone1935vectors}, as a preprocessing step are very common. These methods aim at removing redundant information and lead to more expressive data.

A further problem arises from the fact that the recordings created by cameras of different manufacturers are a priori not compatible. There is no standardization regarding the distribution of the channels in the wavelength space. Therefore, two near-infrared cameras from different manufacturers, which cover the same spectral range, have generally different wavelength assignments for the channels. A model, which identifies the features based on the channel index, will fail on the recordings of another camera. In general, a solution for this problem is standardizing the recording to defined wavelengths. A basic and reliable approach is linear interpolation \cite{steffensen2013interpolation}.

In this work, we want to tackle the two mentioned problems and propose a modified 2D convolution layer optimized for hyperspectral recordings. We outperform comparable approaches and reduce the parameters significantly by inferring a proximity bias for the channel dimension. Further, our model can incorporate the channels' wavelength information. This capability allows the training of camera-agnostic models, meaning the models can perform their tasks on recordings of different cameras. Besides the theoretical background, we prove our claims with empirical experiments on two hyperspectral applications.

Our main contributions are:
\begin{itemize}
    \item We analyze two challenges of hyperspectral recordings. Based on the challenges, we infer a construed bias for hyperspectral models.
    \item A 2D convolution optimized for hyperspectral recording is proposed. The convolution supports camera-agnostic behavior, is interpretable regarding the selected spectral features, and reduces the number of parameters.  
    \item The proposed method is validated in two different hyperspectral applications with publicly available data sets.
\end{itemize}

\begin{figure*}
    \centering
    \begin{subfigure}{0.44 \textwidth}
        \centering
        \includegraphics[height=5.5cm]{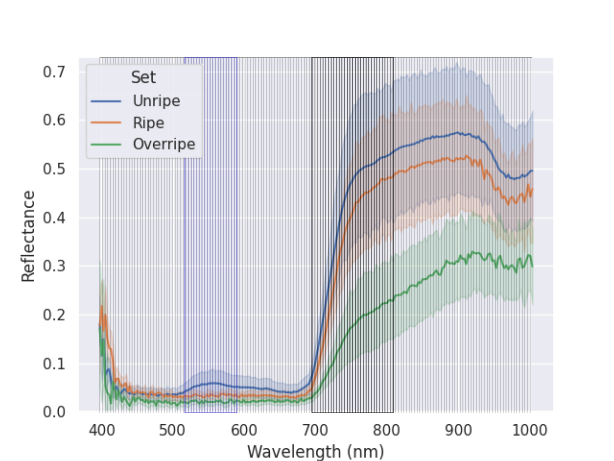}
        \caption{Conv2D}
        \label{fig:avocado_spectra_conv}
    \end{subfigure}
    \begin{subfigure}{0.55 \textwidth}
        \centering
        \includegraphics[height=5.5cm]{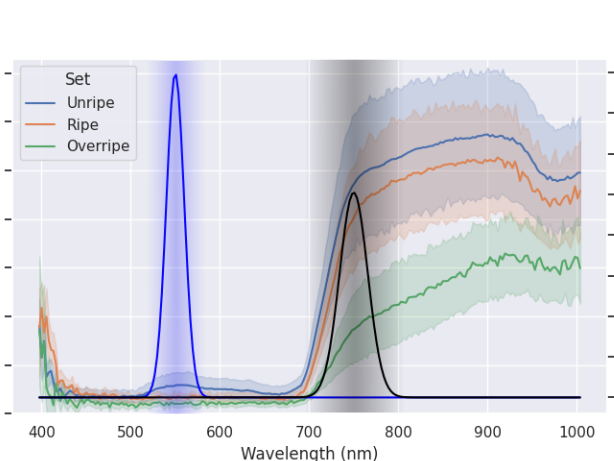}
        \caption{HyveConv (ours)}
        \label{fig:avocado_spectra_ours}
    \end{subfigure}
    \caption{The channel dimension handling visualized for normal 2D convolutions and our approach. Two example features are presented.}
    \label{fig:avocado_spectra_all}
\end{figure*}

\section{Related Work}
\label{sec:related_work}
Classical machine learning approaches, like SVMs \cite{cortes1995support} or k-nearest-neighbors \cite{Mucherino2009}, are still very common for the classification of hyperspectral recordings. In recent years, deep-learning-based methods could outperform these in many hyperspectral applications. Chen \etal were one of the earliest adopters of deep learning for hyperspectral recordings \cite{6844831}. Their approach was based on a PCA followed by stacked autoencoders and a final logistic regression. More recent methods utilize convolution layers and are therefore called convolutional neural networks. These can incorporate the spatial information of the neighboring pixels beside the spectrum of a single pixel. This additional information and the higher complexity of the methods supported their breakthrough. In this work, we focus on convolutional neural networks.

Convolutional neural networks can be divided into methods based on 2D convolutions, 3D convolutions, or a mixture of both. 
2D convolutions perform only spatial convolutions. So the exchange of information between channels is limited and primarily conducted by the final fully connected layers. Makantasis \etal were the first who utilized 2D convolutions for hyperspectral recordings \cite{Makantasis2015DeepNetworks}. 2D convolutions are still very common for hyperspectral recordings\cite{DBLP:conf/ijcnn/VargaMZ21}, because the required number of parameters is limited, and they are still mighty.

In contrast, 3D convolutions can perform convolutions in all three dimensions of the hyperspectral cube. So they can incorporate additional information, but are also parameter hungry. Large models are hard to train on the small hyperspectral data sets. Therefore, many approaches try to optimize the model power-size-ratio. Smaller models with the same performance are preferred because they tend to overfit less and often produce more stable results over different training runs. Hamida \etal used 3D convolutions to classify hyperspectral remote sensing data \cite{8344565}. He \etal introduced a multiscale 3D convolutional neural network, which applies 3D convolutions with different kernel sizes. This boosts the performance of the 3D convolutions. Roy \etal proposed Fusenet, which fuses the output of 3D convolutions by using residual fuse blocks \cite{DBLP:journals/lgrs/RoyKDC20}.

The third category combines 3D convolutions and 2D convolutions. Roy \etal proposed HybdriSN \cite{DBLP:journals/lgrs/RoyKDC20}. This model has a 3D convolution backbone. The output of this backbone is processed by a 2D convolution and a fully connected head.

SpectralNET \cite{chakraborty2021spectralnet} belongs to the first category, but it mimics 3D convolutions with wavelet transformations; therefore, it is also part of the third category.

Vision transformers, the most recent computer vision trend, also impacted the HSI classification. There are already some adaptations for hyperspectral recordings \cite{DBLP:journals/remotesensing/QingLFG21a, DBLP:journals/tgrs/HongHYGZPC22}. But for this application, the transformer models can often only perform comparably performance to convolutional neural networks. Therefore, the smaller convolutional neural networks are often preferred.

Our method is based on 2D convolutions and, therefore, part of the first group. We utilize the wavelength meta information of the input channels to learn a continuous representation of the features in the input channel dimension. As our approach only affects the first convolution layer, it is compatible with other works mentioned.

Our method utilizes Gaussian distributions to represent the feature distribution in the input channel dimension. Still, our approach is not related to Bayesian convolutional neural networks \cite{DBLP:journals/corr/abs-1901-02731}. These networks try to approximate the true posterior and incorporate the uncertainty into the inference process, which is not part of our approach.

Hu \etal \ proposed a similar approach \cite{DBLP:journals/pami/HuSASW20}. Their Squeeze-and-Excitation block allows the network to learn a channel interdependency. Our approach differs in three key points. First, their method uses the input to predict weightings for each feature channel. Our method uses meta information, the channels' wavelength, to weigh the kernels.
Further, our convolution introduced the bias that channels with similar wavelengths should use similar kernels. This proximity relation is helpful for hyperspectral records, shown in section \ref{sec:motivation}. Last, our method also allows the interpretation of the learned features. The selected wavelengths can be visualized and analyzed, as shown in section \ref{sec:learned_wroi}.

Both methods share the idea of introducing an interdependency in the channel dimension. In the experiments, we can also show that our approach outperforms their approach in the hyperspectral application.

\section{Proposed Method}
The approach is based on 2D convolutions. Regular 2D convolutions handle input based on the input channel, which is not optimal for hyperspectral recordings. We emphasize key problems and justify our modifications. Further, we propose the method itself. The procedure is evaluated in the section afterward with experiments on two hyperspectral data sets with different applications.

\subsection{Motivation}
\label{sec:motivation}
The reflected light, recorded by a camera, is a spectrum of many wavelengths. An RGB image oversimplifies this spectrum with three sampling bands (red, green, and blue). Hyperspectral recordings record many more bands and mimic a much better spectrum approximation. Fig. \ref{fig:avocado_spectra_conv} shows the spectra of avocados with different ripening states recorded with HSI. The continuity of the underlying spectrum is captured sufficiently.

As a result, we encounter the two problems mentioned in the introduction. First, the numerous channels of the input data (around 200 bands) would demand a large first convolution layer. For the proposed method, the network should have access to the entire hyperspectral cube without a dimension reduction as preprocessing. A dimension reduction could reduce the size of the hyperspectral cube. But by using a dimension reduction, the original data can only be approximated. We argue that if the network can use the full potential of the data, this usually is beneficial for the selected features, as deep learning approaches can handle high dimensional data well \cite{doi:10.1126/science.1127647}. We prove this in the experiment section empirically.

The second problem is that the recorded wavelengths of hyperspectral cameras are not standardized. Recordings of a manufacturer's near-infrared (NIR) camera are usually not compatible with recordings of a NIR camera of another manufacturer, even though both cameras share the same wavelength range. Their channel-wavelength assignments are often shifted and have different gradients. An example can be found in the supplementary material. As 2d-convolutions are based on the index of channels, standardizing the data by a preprocessing step, like linear interpolation\ \cite{davis1975interpolation}, is necessary. These preprocessing steps harm the end-to-end training. We propose a method capable of handling different hyperspectral cameras by design.

To solve the mentioned problems, our convolution learns a wavelength range of interest (WROI) for each feature instead of the specific input channel. By having a continuous representation of the channel dimension, it can sample the kernels based on the wavelengths of the input channels. 

By adding the bias, that the network should apply similar kernels for similar wavelengths, it is possible to reduce the number of parameters significantly. This bias restricts the freedom of the model. Still, in the context of a continuous spectrum in the channel dimension, this is reasonable and seems one key point for handling hyperspectral recordings.

In summary, we propose a method that adds a bias regarding neighboring channels. It eliminates the need for dimensionality reduction for hyperspectral images. And it enables the training of hyperspectral camera-independent models. In section \ref{sec:experiments}, we provide empirical evidence for these claims. But first, the method itself is described. 

\begin{figure}
    \centering
    \includegraphics[width=\columnwidth]{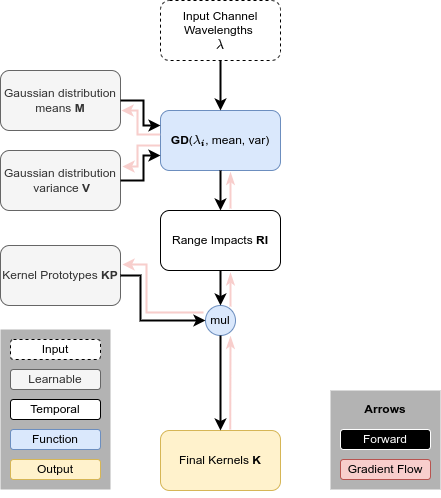}
    \caption{Flow of Hyperspectral Visual Embedding Convolution (HyveConv)}
    \label{fig:hyveconv_flow}
\end{figure}

\subsection{Method}
\label{sec:method}
The fundamental idea of this approach is to learn kernels and their target wavelength range in combination  (Fig. \ref{fig:avocado_spectra_ours}) instead of learning each input channel kernel independently (Fig. \ref{fig:avocado_spectra_conv}). A learnable Gaussian distribution represents a wavelength range. The weighting factor for the corresponding kernel for this input channel is given by sampling the distribution at the input channel wavelength. The resulting kernel is then calculated by multiplying the factor and the kernel. Finally, the kernel is used for a 2D convolution on the specific input channel. Fig. \ref{fig:hyveconv_flow} shows the method.

In the following, the method is described in further detail. Afterward, an extension is explained, which adds more synergy effects for the kernels. 


A 2D convolution calculates the cross-correlation between trainable kernels and the input data. For $C_{in}$ input channels, $C_{out}$ output channels and kernel size of $K_x \cdot K_y$, this results in a matrix $W$ for the trainable weights:

\begin{equation}
    W \in \mathbb{R}^{C_{in} \times C_{out} \times K_x \times K_y}
    \label{eq:W}
\end{equation}

The number of trainable parameters of a convolution depends on the number of input channels. For the first layer, the input channels are defined by the channel dimension of the input data. For hyperspectral recordings, this is around 200. For depthwise-separable convolutions \cite{DBLP:conf/cvpr/Chollet17} this relation is weakened but still exists (further analysis in the supplementary material).

To tackle the issue of a too large first layer, we learn wavelength ranges of interest (WROIs) for the kernels instead of channel-wise kernels. We assume that adjacent channels of the wavelength space typically share similar behavior. An example of this behavior can be found in Fig. \ref{fig:avocado_spectra_conv}, where adjacent channels have very similar reflectance values, originating from the high resolution in the wavelength dimension and the continuity of the signal. Both points can be expected for HSI, so the bias to use similar kernels in neighboring bands seems suitable and even crucial.

Our convolution decouples the number of kernels from the number of input channels. Instead of kernels per input channel, wavelength ranges of interest (WROI) and their kernels are learned. $G$ defines the number of WROIs. Their learnable kernels are called kernel prototypes (KP). This results in the matrix for the kernel prototype weights:
\begin{equation}
    KP \in \mathbb{R}^{G \times C_{out} \times K_x \times K_y}
    \label{eq:W_kernel}
\end{equation}
For the learnable distributions, Gaussian distributions (GDs) are used. A Gaussian distribution (GD) can mimic the wanted behavior, that the impact decays to the borders of a WROI. Further, it is defined by just two parameters, the mean $\mu$ and the variance $\sigma^2$. Both parameters are differentiable and interpretable. A Gaussian distribution is the combination of a learnable mean $\mu$ and a learnable variance $\sigma^2$. The value of the Gaussian distribution for a value $x$ is defined as  Eq. \ref{eq:Gaussian}.
\begin{equation}
    GD(x, \mu, \sigma^2) = \frac{1}{\sqrt{2\pi\sigma^2}} \exp\left(-\frac{x-\mu}{2\sigma^2}\right)
    \label{eq:Gaussian}
\end{equation}

To predict the kernels for specific channel wavelengths $\lambda$, the first step is to calculate the Gaussian distributions at these wavelengths $\lambda$. The result is the range impact matrix $RI$, which defines the impact of all kernel prototypes on all input channels:
\begin{equation}
\begin{aligned}
    &RI \in \mathbb{R}^{C_{in} \times G} \\
    &\text{with } RI_{ij} = GD(\lambda_i, \mu_j, \sigma^2_j)
    \label{eq:RI}
\end{aligned}
\end{equation}
Afterward, the learnable kernel prototypes can be weighted with this matrix to produce the final kernels $K$. These kernels are then used for a 2D convolution on the input.
\begin{equation}
    K = RI \cdot KP \in \mathbb{R}^{C_{in} \times C_{out} \times K_x \times K_y}
    \label{eq:K}
\end{equation}
The result of this convolution is input to further layers.
Our convolution can reduce the trainable parameters to $W_{hyve}$ with $G << C_{in}$. In Tab. \ref{tab:trainable_parameters} the number of trainable parameters of the different models is compared.
\begin{equation}
    W_{hyve} = \left( KP, M \in \mathbb{R}^G, V \in \mathbb{R}_{>0}^G \right)
    \label{eq:W_our}
\end{equation}
To support full end-to-end learning, a gradient for Gaussian distributions and kernel prototypes is needed. Fig. \ref{fig:hyveconv_flow} shows the gradient flow in our convolution model. The multiplication divides the gradient of the final kernel $K$ on the learnable kernel prototypes $KP$ and the range impact matrix $RI$. The matrix $RI$ holds entries, which were sampled, of the Gaussian distributions regarding the wavelengths of the input. This allows us to infer the impact of the input wavelengths on the gradient. Further, the gradient can be passed to the means $M$ and variances $V$ of the Gaussian distributions. The channel wavelengths $\lambda$ are part of the input data and do not need a gradient. So, all learnable components of the convolution are trainable based on the gradient of the final kernel $K$, and end-to-end training of the model is possible.

In the supplementary material, the behavior of the introduced hyperparameter $G$ is discussed in further detail.

\begin{table}[t]
\caption{Parameters of a single convolution for the configuration: $C_{in}=200$, $C_{out}=25$, $K_{x}=3$, $K_{y}=3$, $G=5$}
\label{tab:trainable_parameters}
\resizebox{0.95\columnwidth}{!}{%
\begin{tabular}{l|l|l|l}
\textbf{Conv2D} & \textbf{\begin{tabular}[c]{@{}l@{}}Depthwise\\ -separable\\ Conv2D\end{tabular}} & \textbf{\begin{tabular}[c]{@{}l@{}}HyveConv \\ (ours)\end{tabular}} & \textbf{\begin{tabular}[c]{@{}l@{}}HyveConv++\\ (ours)\end{tabular}} \\ \hline
\begin{tabular}[c]{@{}l@{}}$200 \cdot 25 \cdot 3 \cdot 3$ \\ $= 45000$\end{tabular} & \begin{tabular}[c]{@{}l@{}}$200 \cdot 1 \cdot 3 \cdot 3$\\ $+ 200 \cdot 25  \cdot 1 \cdot 1$\\ $= 6800$\end{tabular} & \begin{tabular}[c]{@{}l@{}}$5 \cdot 25 \cdot 3 \cdot 3$\\ $+ 2 \cdot 5$\\ $= \textbf{1135}$\end{tabular} & \begin{tabular}[c]{@{}l@{}}$5 \cdot 25 \cdot 3 \cdot 3$\\ $+ 2 \cdot 5$\\ $+ 1 \cdot 25 \cdot 3 \cdot 3$\\ $+ 1 \cdot 1 \cdot 3 \cdot 3$\\ $+ 2$\\ $= \textbf{1371}$\end{tabular}
\end{tabular}
}
\end{table}

\subsubsection{Extension: Additional Kernel Sharing}
The learned WROIs allow the model to share kernels through the channel dimension of the input $C_{in}$.\\
As an extension, we propose sharing parts of kernels through the channel dimension of the output $C_{out}$ and overall kernels of the convolution layer. For this, the previous method is enhanced with additional kernel prototypes. These additional kernel weights are weighted with the learnable factors $\alpha \in \mathbb{R}$ and $\beta \in \mathbb{R}$. The sum of all kernel prototypes is then used further.
\begin{equation}
\begin{aligned}
    &KP_{++}^{i,j,m,n} = KP^{i,j,m,n} + \alpha \cdot KP_{c\_out}^{1,j,m,n} + \beta \cdot KP_{conv}^{1,1,m,n}\\
    \label{eq:W_kernel++}
    &KP_{c\_out} \in \mathbb{R}^{1 \times C_{out} \times K_x \times K_y}\\
    &KP_{conv} \in \mathbb{R}^{1 \times 1 \times K_x \times K_y}
\end{aligned}
\end{equation}
The kernel prototypes $KP_{++}$ replace the kernel prototypes $KP$ in Eq. \ref{eq:K} resulting in Eq. \ref{eq:K++}, which predicts the final kernels.
\begin{equation}
    K_{++} = RI \cdot KP_{++} \in \mathbb{R}^{C_{in} \times C_{out} \times K_x \times K_y}
    \label{eq:K++}
\end{equation}
With this extension, our approach has the following trainable parameters:
\begin{equation}
    W_{hyve++} = \left( KP, M, V, \alpha, KP_{c\_out}, \beta, KP_{conv} \right)
    \label{eq:W_our++}
\end{equation}
With this extension, the model can utilize the synergy effects within the kernels of a convolution layer better.

It is important to keep the impact of the shared kernel prototypes at the beginning small. Otherwise, the training is very unstable. Therefore, we recommend an initial value for $\alpha_0$ and $\beta_0$ of $0.1$. An evaluation of the impact of the proposed extension can be found in the supplementary material.

\section{Experiments}
\label{sec:experiments}
The proposed method is evaluated in two hyperspectral applications in the following section. The first application covers a classification task of ripening fruit recorded under laboratory conditions \cite{DBLP:conf/ijcnn/VargaMZ21}. The second covers a well-established segmentation task of remote sensing data recorded by satellites.

For the following experiments, the extended version of the proposed method (see Eq. \ref{eq:K++}) with the following parameters was used: $G=5$, $\alpha_0=0.1$ and $\beta_0=0.1$. Each composition was tested with three random seeds. The random seed affects the network initialization, the training sample order, and the data augmentation order. 

\begin{figure}
    \centering
    \includegraphics[width=\columnwidth]{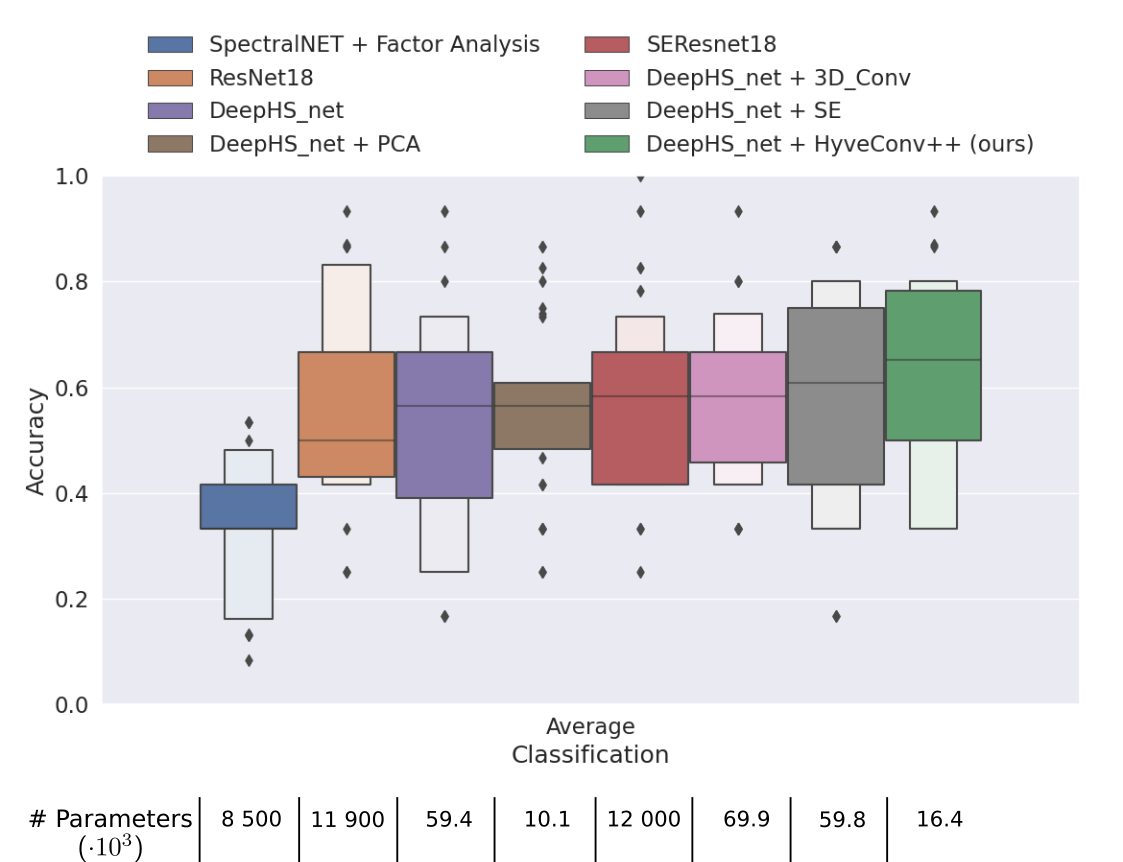}
    \caption{Overall accuracy on the ripening fruit data set with the Specim FX 10 recordings.}
    \label{fig:deephs_results}
\end{figure}

\subsection{Application A: Fruit Ripeness Prediction}
The first application's task is to classify fruit's ripeness level. The used data set \cite{DBLP:conf/ijcnn/VargaMZ21} for this application is one of the largest labeled hyperspectral data sets publicly available. In addition, there are nearly no data set with recordings of different hyperspectral cameras for the same scene available. This property is relevant for testing the camera-agnostic behavior of our model.
\paragraph{Data Set}
The data set covers four fruit types (avocado, kiwi, persimmon, and mangos). Ripeness is classified into three categories (firmness, sweetness, and overall ripeness). Avocados have no sweetness; therefore, these only cover the first and last categories. This sums up to eleven different setups (e.g., the firmness prediction for mangos or the sweetness prediction for kiwis).
All setups cover three classes (unripe, perfect, and overripe). This data set has fixed training and test sets.
Most of the recordings were done by a Specim FX 10. This hyperspectral camera covers the wavelength range of $397.66 nm$ to $1003.81 nm$ with 224 bands. In addition, there are many recordings of a Corning microHSI 410 Vis-NIR Hyperspectral Sensor. It covers the wavelengths between $408.03 nm$ and $901.26 nm$ with 249 bands. All recordings are already normalized with a white and a dark reference.

We use the training and test pipeline proposed in \cite{DBLP:conf/ijcnn/VargaMZ21}.

\paragraph{Models} For the experiments, we used four models.
As a basis for our approach, the model DeepHS\_net\cite{DBLP:conf/ijcnn/VargaMZ21} is used. It is a shallow convolutional neural network consisting of three depthwise-separable convolutions, a global average pooling layer, and a fully connected head. The complexity of the model is low, which helps in understanding the model internals. Further, the model could already achieve satisfying results for the prediction of the ripeness level of fruit. It is optimized for the small size of hyperspectral data sets.

For our model, we replaced the first convolution layer with a \convname{}++ layer, keeping all dimensions of the convolution fixed. The rest of the model stays the same. The training schedule of the original publication is used.

Further, we used a ResNet18 \cite{He2016DeepRecognition}, which is still a commonly used backbone, and SpectralNET with factor analysis \cite{chakraborty2021spectralnet}, which achieves state-of-the-art performance on the remote sensing data set. Finally, we used a Squeeze-and-Excitation (SE) \cite{DBLP:journals/pami/HuSASW20} in two variants. First, in combination with a ResNet18 network and second with the DeepHS\_net model. The SE block adds interdependency in the channel dimension of the kernels, which is similar to our approach and therefore interesting for comparison.

\begin{figure}
\centering
    \includegraphics[width=\columnwidth]{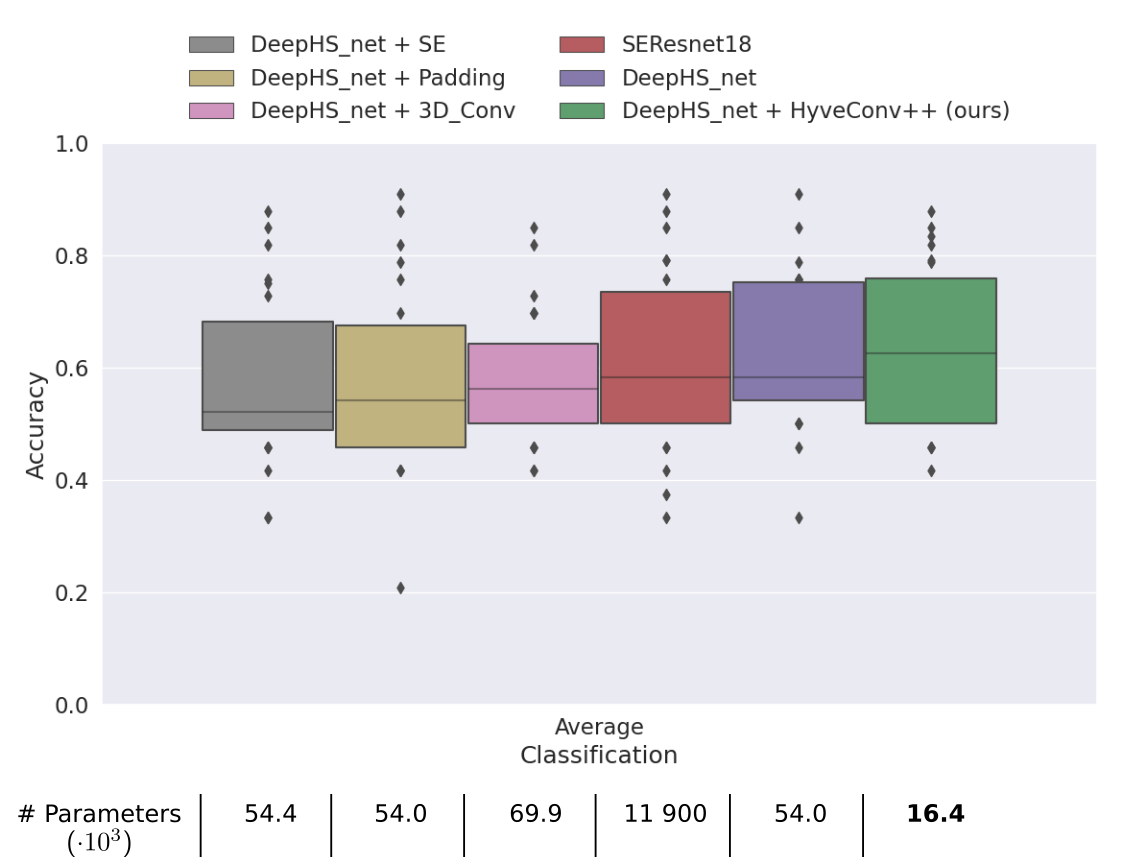}
    \caption{Overall accuracy on the ripening fruit data set with recordings of both cameras.}
    \label{fig:deephs_results_real_setup}
\end{figure}

\begin{table*}[]
\centering
\caption{Classification accuracies (\%) of different models in terms of OA, Kappa, and AA with varying training data 10\% and 30\%, respectively. Based on the evaluations of Chakraborty \etal \cite{chakraborty2021spectralnet}. Two configuration of our model are presented. (*) not fine-tuned (**) larger hidden layers.}
    \label{tab:resuls_hrss}
\resizebox{0.95\textwidth}{!}{%
\begin{tabular}{c|cr|ccc|ccc|ccc}
\textbf{Training} & \textbf{Methods} & \multicolumn{1}{l|}{\textbf{\# of params}} & \textbf{} & Indian Pines dataset &  &  & Pavia University dataset &  &  & Salinas dataset &  \\
\textbf{Samples} &  & \multicolumn{1}{l|}{} & OA & Kappa & AA & OA & Kappa & AA & OA & Kappa & AA \\ \hline
\multirow{9}{*}{10\%} & SVM \cite{cortes1995support} &  & 81.67$\pm$0.7 & 78.76$\pm$0.8 & 79.84$\pm$3.4 & 90.58$\pm$0.5 & 87.21$\pm$0.7 & 92.99$\pm$0.4 & 94.46$\pm$0.1 & 93.13$\pm$0.3 & 93.01$\pm$0.6 \\
 & 2D-CNN \cite{7326945} & 561,300 & 80.27$\pm$1.2 & 78.26$\pm$2.1 & 68.32$\pm$4.1 & 96.63$\pm$0.2 & 95.53$\pm$0.2 & 94.84$\pm$1.4 & 96.34$\pm$0.3 & 95.93$\pm$0.9 & 94.36$\pm$0.5 \\
 & 3D-CNN \cite{8344565} & 991,596 & 82.62$\pm$0.1 & 79.25$\pm$0.3 & 76.51$\pm$0.1 & 96.34$\pm$0.2 & 94.90$\pm$1.2 & 97.03$\pm$0.6 & 85.00$\pm$0.1 & 83.20$\pm$0.7 & 89.63$\pm$0.2 \\
 & M3D-CNN \cite{8297014} & 372,544 & 81.39$\pm$2.6 & 81.20$\pm$2.0 & 75.22$\pm$0.7 & 95.95$\pm$0.6 & 93.40$\pm$0.4 & 97.52$\pm$1.0 & 94.20$\pm$0.8 & 93.61$\pm$0.3 & 96.66$\pm$0.5 \\
 & FuSENet \cite{DBLP:journals/iet-ipr/RoyDCC20} & 100,880 & 97.11$\pm$0.2 & 97.25$\pm$0.2 & 97.32$\pm$0.2 & 97.65$\pm$0.3 & 97.69$\pm$0.3 & 97.68$\pm$0.4 & 99.23$\pm$0.1 & 99.97$\pm$0.2 & 99.16$\pm$0.1 \\
 & HybridSN \cite{DBLP:journals/lgrs/RoyKDC20} & 5,122,176 & 98.39$\pm$0.1 & 98.16$\pm$0.1 & 98.01$\pm$0.2 & \textbf{99.72$\pm$0.1} & \textbf{99.64$\pm$0.1} & 99.20$\pm$0.1 & \textbf{99.98$\pm$0.2} & \textbf{99.98$\pm$0.2} & \textbf{99.98$\pm$0.1} \\
 & SpectralNET \cite{chakraborty2021spectralnet} & 6,800,336 & \textbf{98.76$\pm$0.2} & 98.59$\pm$0.1 & 98.61$\pm$0.1 & 99.71$\pm$0.1 & 99.62$\pm$0.1 & 99.43$\pm$0.2 & 99.96$\pm$0.2 & 99.96$\pm$0.1 & 99.97$\pm$0.1 \\
 & \cellcolor[HTML]{E0E0E0}HyveConv++ (ours) * & \cellcolor[HTML]{E0E0E0}\textbf{16,700} & \cellcolor[HTML]{E0E0E0}98.18$\pm$0.6 & \cellcolor[HTML]{E0E0E0}98.41$\pm$0.1 & \cellcolor[HTML]{E0E0E0}98.28$\pm$0.1 & \cellcolor[HTML]{E0E0E0}99.30$\pm$0.3 & \cellcolor[HTML]{E0E0E0}99.30$\pm$0.3 & \cellcolor[HTML]{E0E0E0}\textbf{99.49$\pm$0.2} & \cellcolor[HTML]{E0E0E0}99.24$\pm$0.2 & \cellcolor[HTML]{E0E0E0}99.64$\pm$0.0 & \cellcolor[HTML]{E0E0E0}99.94$\pm$0.0 \\
 & \cellcolor[HTML]{E0E0E0}HyveConv++ (ours) ** & \cellcolor[HTML]{E0E0E0}\textbf{25,200} & \cellcolor[HTML]{E0E0E0}98.33$\pm$0.1 & \cellcolor[HTML]{E0E0E0}\textbf{98.64$\pm$0.1} & \cellcolor[HTML]{E0E0E0}\textbf{98.69$\pm$0.1} & \cellcolor[HTML]{E0E0E0}99.42$\pm$0.2 & \cellcolor[HTML]{E0E0E0}99.49$\pm$0.2 & \cellcolor[HTML]{E0E0E0}99.46$\pm$0.2 & \cellcolor[HTML]{E0E0E0}99.89$\pm$0.0 & \cellcolor[HTML]{E0E0E0}99.79$\pm$0.0 & \cellcolor[HTML]{E0E0E0}99.74$\pm$0.0 \\ \hline
\multirow{10}{*}{30\%} & SVM \cite{cortes1995support} &  & 87.24$\pm$0.4 & 85.27$\pm$0.5 & 85.15$\pm$1.1 & 95.65$\pm$0.1 & 94.63$\pm$0.2 & 94.60$\pm$0.1 & 94.95$\pm$0.1 & 94.48$\pm$0.1 & 97.93$\pm$0.1 \\
 & 2D-CNN \cite{7326945} & 561,300 & 88.90$\pm$1.3 & 87.01$\pm$1.6 & 85.70$\pm$1.0 & 96.50$\pm$0.4 & 96.55$\pm$0.3 & 96.00$\pm$0.1 & 96.75$\pm$0.6 & 96.71$\pm$0.7 & 98.57$\pm$0.2 \\
 & 3D-CNN \cite{8344565} & 991,596 & 90.23$\pm$0.2 & 89.70$\pm$0.3 & 89.87$\pm$0.1 & 97.90$\pm$0.3 & 97.22$\pm$0.1 & 97.30$\pm$0.1 & 95.54$\pm$0.5 & 94.81$\pm$0.3 & 97.09$\pm$0.6 \\
 & M3D-CNN \cite{8297014} & 372,544 & 95.67$\pm$0.1 & 94.70$\pm$0.3 & 94.60$\pm$0.6 & 97.60$\pm$0.2 & 96.50$\pm$0.6 & 98.00$\pm$0.1 & 94.99$\pm$0.3 & 95.40$\pm$0.1 & 96.28$\pm$0.2 \\
 & FuSENet \cite{DBLP:journals/iet-ipr/RoyDCC20} & 100,880 & 99.01$\pm$0.2 & 98.60$\pm$0.1 & 98.64$\pm$0.1 & 99.42$\pm$0.2 & 99.21$\pm$0.3 & 99.33$\pm$0.2 & 99.68$\pm$0.2 & 99.74$\pm$0.1 & 99.69$\pm$0.1 \\
 & \multicolumn{1}{l}{ImprovedTransformerNet \cite{ DBLP:journals/remotesensing/QingLFG21a}} & 150,000,000 & 99.22 \% & 99.19 \% & 99.08 \% & 99.64 \% & 99.49 \% & 99.67 \% & 99.91 \% & 99.78 \% & 99.63 \% \\
 & HybridSN \cite{DBLP:journals/lgrs/RoyKDC20} & 5,122,176 & 99.75$\pm$0.1 & 99.71$\pm$0.1 & 99.63$\pm$0.2 & 99.98$\pm$0.1 & \textbf{99.98$\pm$0.2} & 99.97$\pm$0.2 & \textbf{100} & \textbf{100} & \textbf{100} \\
 & SpectralNET \cite{chakraborty2021spectralnet} & 6,800,336 & \textbf{99.86$\pm$0.2} & \textbf{99.84$\pm$0.2} & \textbf{99.98$\pm$0.1} & \textbf{99.99$\pm$0.1} & \textbf{99.98$\pm$0.1} & \textbf{99.98$\pm$0.1} & \textbf{100} & \textbf{100} & \textbf{100} \\
 & \cellcolor[HTML]{E0E0E0}HyveConv++ (ours) * & \cellcolor[HTML]{E0E0E0}\textbf{16,700} & \cellcolor[HTML]{E0E0E0}99.85$\pm$0.0 & \cellcolor[HTML]{E0E0E0}99.75$\pm$0.0 & \cellcolor[HTML]{E0E0E0}99.7$\pm$0.2 & \cellcolor[HTML]{E0E0E0}99.97$\pm$0.0 & \cellcolor[HTML]{E0E0E0}99.96$\pm$0.0 & \cellcolor[HTML]{E0E0E0}99.97$\pm$0.0 & \cellcolor[HTML]{E0E0E0}99.98$\pm$0.0 & \cellcolor[HTML]{E0E0E0}99.99$\pm$0.0 & \cellcolor[HTML]{E0E0E0}99.99$\pm$0.0 \\
 & \cellcolor[HTML]{E0E0E0}HyveConv++ (ours) ** & \cellcolor[HTML]{E0E0E0}\textbf{25,200} & \cellcolor[HTML]{E0E0E0}\textbf{99.86$\pm$0.0} & \cellcolor[HTML]{E0E0E0}\textbf{99.84$\pm$0.0} & \cellcolor[HTML]{E0E0E0}99.57$\pm$0.1 & \cellcolor[HTML]{E0E0E0}99.96$\pm$0.0 & \cellcolor[HTML]{E0E0E0}\textbf{99.98$\pm$0.0} & \cellcolor[HTML]{E0E0E0}99.94$\pm$0.0 & \cellcolor[HTML]{E0E0E0}99.93$\pm$0.0 & \cellcolor[HTML]{E0E0E0}99.92$\pm$0.0 & \cellcolor[HTML]{E0E0E0}99.99$\pm$0.0 \\ \cline{1-12} 
\end{tabular}
}
\end{table*}

    

\paragraph{One Camera} In the first experiments, the recordings of the Specim FX 10 were used only. These experiments indicate the performance of the tested models in the default single-camera use case. The results are visible in Fig. \ref{fig:deephs_results}. Our model could outperform the other models' average accuracy, given the median over all categories and fruit types. A DeepHS\_net with Squeeze-and-Excitation blocks (DeepHS\_net + SE) performed second best. Both approaches introduce an interdependency between the channels, which seems helpful. Our model bias that similar wavelengths should have similar features further boosts performance. As a result, our model produced better results with fewer parameters.

\paragraph{Multiple Cameras}
In the second set of experiments, recordings of two different cameras were used (Specim FX 10 and Corning microHSI 410 Vis-NIR). These cameras differ in the recorded wavelengths. Further analysis can be found in the supplementary material. These experiments validate the camera-agnostic behavior. The number of recordings of each camera is not balanced ($\approx 2:1$), which adds another challenge. In these experiments, the training and test set contained recordings of both cameras.

We compared our method with the best models of the previous experiment. Only the \convname{}++ can process recordings with different wavelength-channel assignments by design. For the other models, preprocessing is necessary. The most basic approach is the usage of zero padding in the channel dimension to produce recordings of the same size. A more advanced method is linear interpolation, which is often the first choice for this problem. For linear interpolation, the configuration is crucial. An inappropriate definition of the quantization steps can harm the expressiveness of the recordings. Linear interpolation was the default preprocessing and was used for most of the models.

The results are visible in Fig. \ref{fig:deephs_results_real_setup}. Our method could outperform the other approaches in the average accuracy and the model size. The other approaches could not produce stable results. Linear interpolation could improve the performance in these experiments. The Squeeze-and-Excitation approach, which performed well in the previous experiment, did not improve the performance of the models here.

By using the wavelength information for the convolution, \convname{}++ can boost the camera-agnostic behavior for HSI.

An additional experiment, which supports this claim, is provided in the supplementary material.


\subsection{Application B: Satellite Imagery Segmentation}
\paragraph{Data Set}
The second application, Hyperspectral Remote Sensing Scenes (HRSS), is a collection of hyperspectral satellite images collected by M. Graña, M.A. Veganzons, and B. Ayerdi. The task is to classify the nature of the ground in different settings. Each scene consists of an image with ground truth labels. The most common scenes are Indian Pines (IP), Pavia University (UP), and Salinas (SA). Each scene is handled separately. We followed the data handling procedure of Chakraborty et al. \cite{chakraborty2021spectralnet}. The segmentation task is converted into a classification task of 64x64 patches.

Indian Pines was recorded with the AVIRIS sensor, which covers the range of $400 nm$ to $2500 nm$ with 224 channels \cite{PURR1947}. Twenty-four noisy channels were already removed from these recordings. The classification happens within 16 classes. Salinas was also recorded by this sensor and covers six classes. Pavia University was recorded with the ROSIS sensor, covering the range from $430 nm$ to $830 nm$ with 103 bands and distinguishing nine classes.

\paragraph{Models}
We used two configurations of our model. Both used the whole hyperspectral cube without dimension reduction. The first configuration (*) is the same model used for the ripening classification task. Only the final output layer was adapted to the number of classes.

The second configuration (**) was slightly adapted for this application. The main change here was to increase the size of the hidden layers, reasoned by the higher number of classes for this application.

\label{sec:results_hrss}
\paragraph{Results} Tab. \ref{tab:resuls_hrss} provides an overview of the performance of the models in the remote sensing application. We tested a variety of different models. The models cover classical machine learning \cite{cortes1995support}, 2D convolutions \cite{7326945}, 3D convolutions \cite{8344565}, mixture of 2D and 3D convolutions \cite{8297014, DBLP:journals/iet-ipr/RoyDCC20}, 2D convolutions with optimized preprocessing \cite{DBLP:journals/lgrs/RoyKDC20, chakraborty2021spectralnet} and one vision transformer based approach \cite{DBLP:journals/remotesensing/QingLFG21a}. Especially, HybridSN \cite{DBLP:journals/lgrs/RoyKDC20} and SpectralNET \cite{chakraborty2021spectralnet} achieve state-of-the-art results for this application. 
The model without modifications (**) could already achieve second/third rank in the overall list (see Tab. \ref{tab:resuls_hrss}). With minor modifications (**), it even achieved state-of-art performance with 250 times fewer parameters. These experiments show that our proposed method generalizes well to different hyperspectral applications.

\begin{figure*}
    \centering
    \begin{subfigure}{0.31 \textwidth}
        \includegraphics[width=\columnwidth]{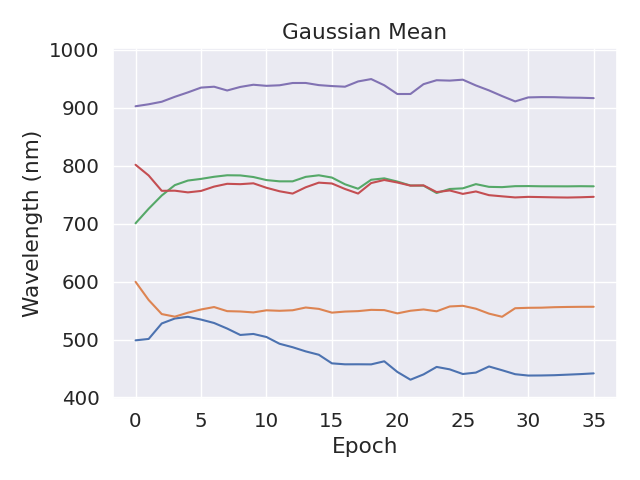}
        \caption{}
        \label{fig:resulting_view_mean}
    \end{subfigure}
    \begin{subfigure}{0.31 \textwidth}
        \includegraphics[width=\columnwidth]{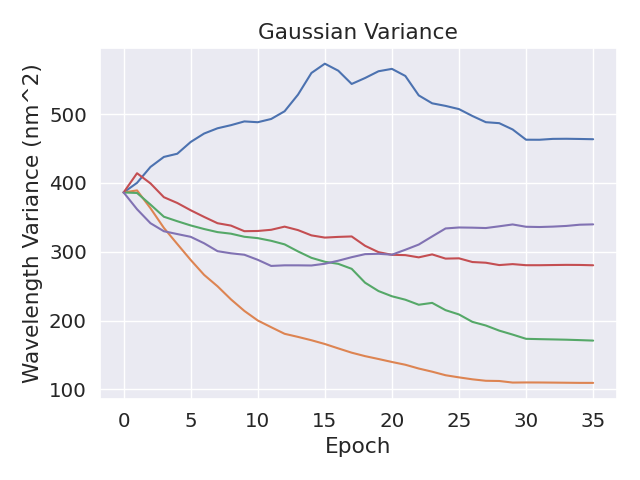}
        \caption{}
        \label{fig:resulting_view_variance}
    \end{subfigure}
    \begin{subfigure}{0.31 \textwidth}
        \includegraphics[width=\columnwidth]{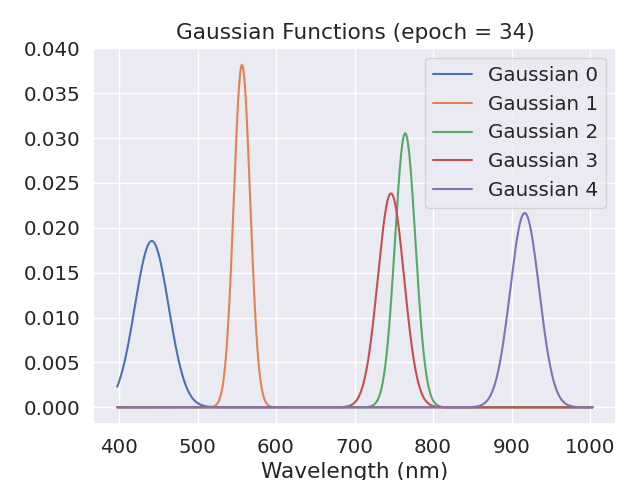}
        \caption{}
        \label{fig:resulting_view_wrois}
    \end{subfigure}
    \begin{subfigure}{\textwidth}
    \centering
    \includegraphics[width=0.55\textwidth]{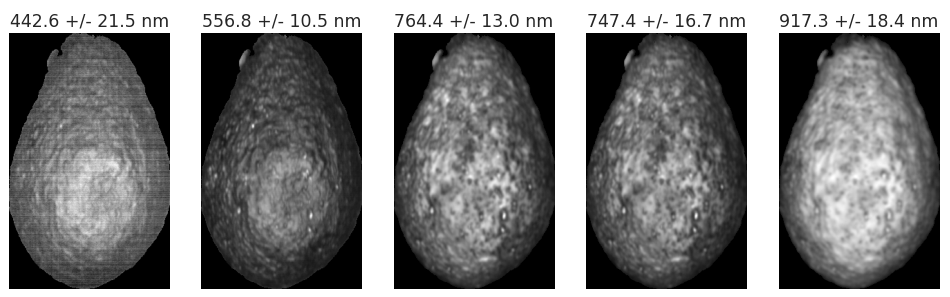}
    \caption{}
    \label{fig:resulting_view_image}
    \end{subfigure}
    \caption{Training of the Gaussian distributions. (a) and (b) show the development of the mean and variance over training epochs. (c) shows the final Gaussian distributions, and these are applied as filters in (d).}
    \label{fig:resulting_view}
\end{figure*}

\section{Ablation Study}
\label{sec:ablation_study}
\label{sec:learned_wroi}
In the previous section, our model outperformed comparable models in two applications. In this section, the interpretability of the learned WROIs, also called camera filters, is presented. For an in-depth ablation study, we refer to the supplementary material. It contains an analysis of the influence of the hyperparameter $G$, defining the number of WROIs. Further, the proposed extension of the method (Eq. \ref{eq:W_our++}) is analyzed. And it is checked whether the training of the Gaussian distribution is necessary. 

Fig. \ref{fig:resulting_view} presents the training and the resulting WROIs for an example run for the ripeness classification of avocados. \ref{fig:resulting_view_mean} and \ref{fig:resulting_view_variance} show the value over epochs for the mean and the variance, respectively. The variance is a good indicator for the WROI search procedure. A decreasing variance indicates that the model found a feature and narrows the corresponding wavelength range. The final WROIs are visible in Fig. \ref{fig:resulting_view_wrois} and the visualizations of these are shown in Fig. \ref{fig:resulting_view_image}. Gaussian distributions 2 and 3 have a final overlap of over $50\%$. They even had a position swap in epoch 3. Therefore, they may cover the same feature, and a reduction of the number of the WROI $G$ could be possible. The final camera filters cover the visible light in the ranges of blue and green. Further, overtones of water and the region, which indicate the degeneration of chlorophyll \cite{WELLBURN1994307}, were selected. These ranges fit the findings of previous works \cite{pinto2019classification, MELADOHERREROS2021111683}. The WROI selection of the trained model could be used to build a multispectral camera optimized for this use case. A multispectral camera with 5-10 custom-defined wavelengths is normally easier to apply in an inline scenario.

We showed that the learned features are explainable and can even be used further.


\section{Conclusion}
In this work, we proposed a 2D convolution optimized for hyperspectral recordings. By using a continuous representation of the kernels and adding a suitable model bias, it is possible to significantly reduce the number of parameters. Further, sampling the kernels by the wavelengths of the input allows the training of a camera-agnostic model. The whole model is end-to-end trainable. And it introduces only one interpretable hyperparameter $G$. The parameter defines the number of wavelength ranges of interest, also called camera filters. Experiments on two different hyperspectral applications could confirm the advantage of this method.

The convolution is proposed for hyperspectral imaging. Still, it could also be helpful in other scenarios of image data with many channels and some proximity relationship between them. This is part of future research.

Multispectral and color cameras are typically based on wavelength filters. The WROIs, learned by our method, are these kinds of filters. Therefore, the model learns the necessary filters for a specific task in a data-driven way. These  could be used to build an optimized multispectral camera.

\ifwacvfinal
\section*{Acknowledgment}
The German Ministry for Economic Affairs and Energy has supported this work, Project Avalon, FKZ: 03SX481B.
The computing cluster of the Training Center Machine Learning, Tübingen has been used for the evaluation, FKZ: 01IS17054.
\fi

\newpage

{\small
\bibliographystyle{ieee_fullname}
\bibliography{egbib}
}

\end{document}


\title{Wavelength-aware 2D Convolutions for Hyperspectral Imaging\\ - Supplementary Material -}

\author{Leon Amadeus Varga, Martin Messmer, Nuri Benbarka, Andreas Zell\\
	Cognitive Systems Group\\
	University of Tuebingen\\
	Tuebingen, Germany\\
	{\tt\small leon.varga@uni-tuebingen.de, martin.messmer@uni-tuebingen.de,}\\ {\tt\small nuri.benbarka@uni-tuebingen.de, andreas.zell@uni-tuebingen.de}}

\maketitle
\thispagestyle{empty}

\section{Introduction}
First, further details about the implementation of the method are given. In section \ref{sec:compare_cameras}, the bands of recordings of the cameras Specim FX 10 and Corning microHSI 410 Vis-NIR are compared. These two cameras were used in the experiments of application A. Then, to strengthen the argument of camera agnostic behavior, we provide a third set of experiments on the fruit set with two simulated cameras. Afterward, common objections are tackled. The supplementary material is rounded out by ablation study experiments that did not find a place in the main paper and a visualization of the prediction on the HRSS data set.


\section{Method}
In this section, we provide additional notes regarding the implementation. Further, we describe the number of learnable parameters for depthwise-separable convolutions in more detail.
\section{Implementation of the Learnable Gaussian}
For our implementation, we initialized the Gaussian distributions of the WROIs in a manner that the whole inspected wavelength range is covered. This is achieved by distributing the means $\mu_{t=0}$ evenly between the minimal ($w_{min}$) and the maximal inspected wavelengths ($w_{max}$). Further, the variance $\sigma^2_{t=0}$ is initialized with overlap:
\begin{equation}
    \sigma^2_{t=0} = \frac{1}{G}^2 \cdot (w_{max} - w_{min})
\end{equation}
Further, negative variances are prevented by using softplus \cite{DBLP:journals/corr/abs-1901-02731}.

\subsection{Depthwise-Separable Convolution}
Depthwise-separable convolutions\cite{DBLP:conf/cvpr/Chollet17} reduce the number of parameters by splitting up a convolution into a spatial- and a channel-based convolution, the overall relation between input channels and necessary weights still exists. Our baseline model DeepHS\_net is based on depthwise-separable convolutions \cite{DBLP:conf/cvpr/Chollet17}. For a depthwise-separable convolution, the learnable parameters are necessary:
\begin{equation}
\begin{aligned}
    W_{ds} = {} &  (W_{spatial} \in \mathbb{R}^{C_{in} \times 1 \times K_x \times K_y}, \\
                &  W_{depth} \in \mathbb{R}^{C_{in} \times C_{out} \times 1 \times 1})
\label{eq:W_ds}        
\end{aligned}
\end{equation}

\begin{figure}
    \centering
    \includegraphics[width=\columnwidth]{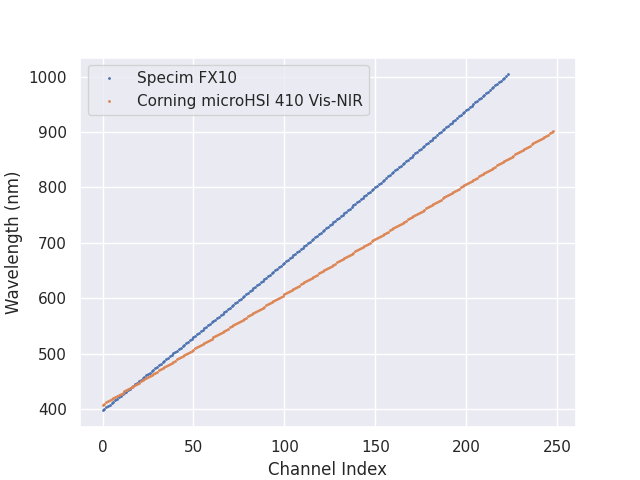}
    \caption{Assigment of wavelengths to channels for Specim FX10 and Corning microHSI 410 Vis-NIR}
    \label{fig:wavelength-channel-assignment}
\end{figure}
\section{Channels of Specim FX10 and Corning MicroHSI 410 Vis-NIR}
\label{sec:compare_cameras}
For our approach, only the difference in the wavelength-channel-assignment is substantial (shown in Fig. \ref{fig:wavelength-channel-assignment}). Other architectural decisions solve the handling of differences in spatial resolution.

For the two selected cameras,  the channels over index 50 differ widely between the two cameras and can result in invalid predictions without adequate handling. Also, the number of total channels slightly differs for the cameras. We show that the proposed method can handle both challenges.

\begin{figure}
    \centering
    \includegraphics[width=\columnwidth]{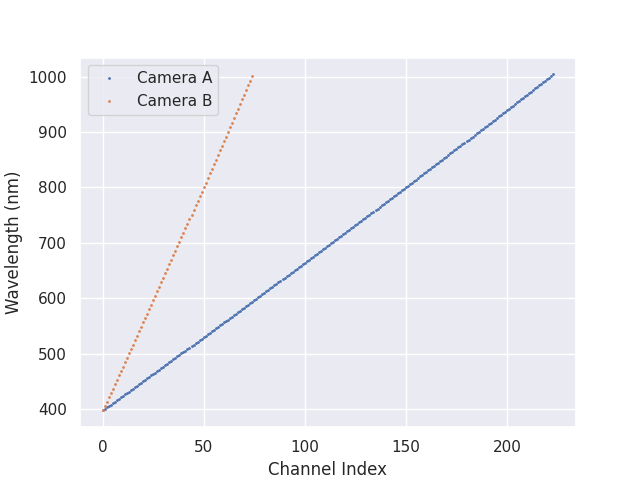}
    \caption{Assignment of wavelengths to channels for the synthetic data}
    \label{fig:wavelength-channel-assignment-synthetic}
\end{figure}

\begin{figure}
\centering
    \includegraphics[width=\columnwidth]{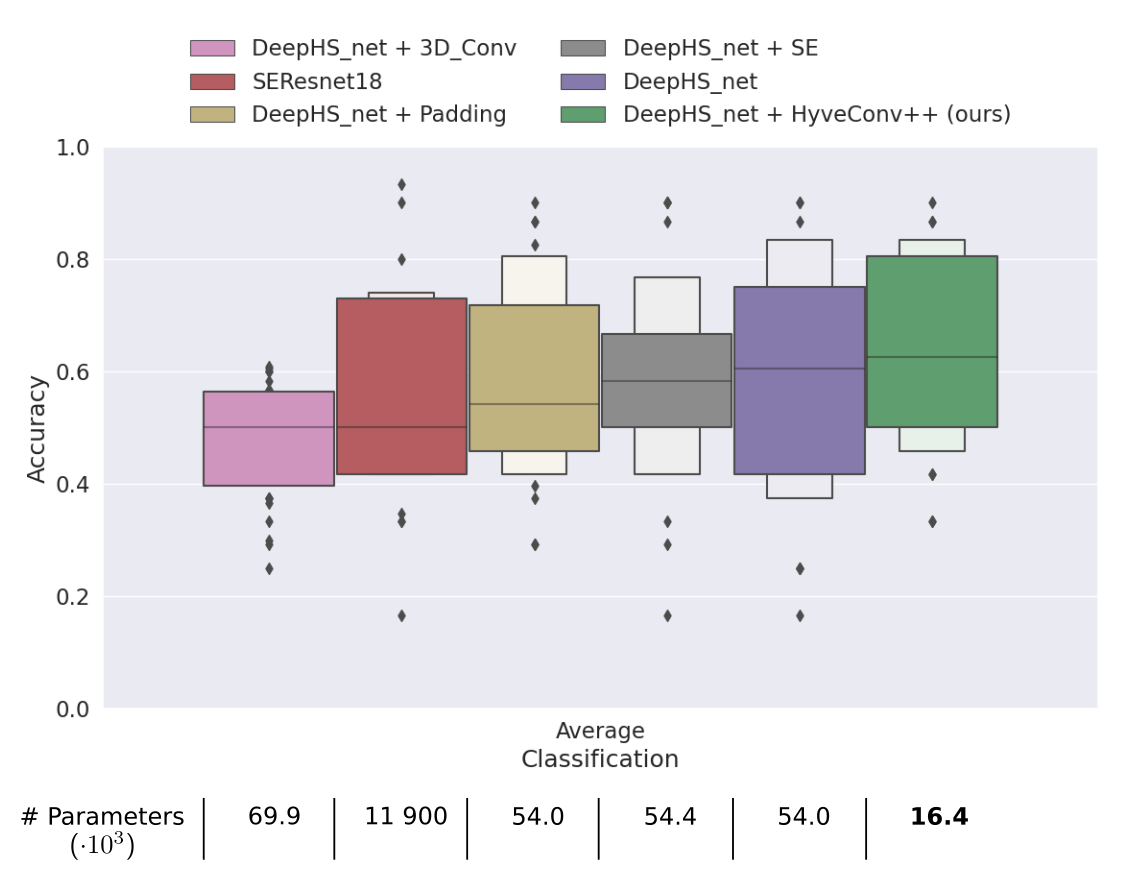}
    \caption{Overall accuracy on the ripening fruit data set with recordings generated for two synthetic cameras.}
    \label{fig:deephs_results_synthetic}
\end{figure}

\section{Experiments for Camera-Agnostic Behavior on Synthetic Data}
There is a lack of hyperspectral data sets with multiple camera recordings of the same scene. We evaluated our approach on a synthetic generated data set to support the claim of camera agnostic behavior. For this, a data set was created based on the Specim FX10 recordings of the ripening fruit \cite{DBLP:conf/ijcnn/VargaMZ21}. Two cameras are simulated using two channels' subsets (shown in Fig. \ref{fig:wavelength-channel-assignment-synthetic}). We selected the channels of the subsets based on different step sizes for the channel indices. This mimics the real setup's behavior, as shown in Fig. \ref{fig:wavelength-channel-assignment}. The wavelengths of the latter channels differ more in contrast to the actual setup.
Further, based on the sampling method of the channels of the two subsets, camera B contains fewer channels than Camera A. Therefore, this experiment setup's challenge is slightly different from the configuration with two real cameras. We used the same models which were used in the main work.

Fig. \ref{fig:deephs_results_synthetic} shows the final outcome of the experiments. Our method could again outperform the other approaches. For these experiments, the baseline model DeepHS\_net with a linear interpolation could produce satisfying results, too. As already mentioned, the challenges of this experiment setup slightly differ from a real camera setup. But we could show that our model could handle both situations better than the other tested models.

\section{Common Objections}
Here we want to tackle the most common objections.
\paragraph{Additional hyperparameter}
The parameter $G$ has a significant role in the method. But it is interpretable, and the default value was stable for two hyperspectral applications. We analyze the stability of the parameter in the ablation study (see section \ref{sec:abl_g}). Further, a basic understanding of the used wavelength range is necessary for hyperspectral application. Thus, a rough intuition of how many WROIs are required for a task is expectable.
Further, the learned WROIs can be visualized, and it is possible to interpret them. For example, $G$ can be increased or decreased based on their final overlap.
\paragraph{Slower inference} The inference time is not affected by this modification. For the training, the prediction of the final kernel $K$ is required in each step. For the inference, this can be done once per camera. No trainable part of the convolution is changing anymore. So the only difference in the inference time is the one-time prediction of the kernels for a new camera type.
\paragraph{Training stability} The training was stable for two significantly different applications in our experiments. Also, the selection of the WROI worked reliably in these settings.

\section{Ablation Study}
This section analyzes the impact of a couple of design choices.
\begin{figure}
    \centering
    \includegraphics[width=\columnwidth]{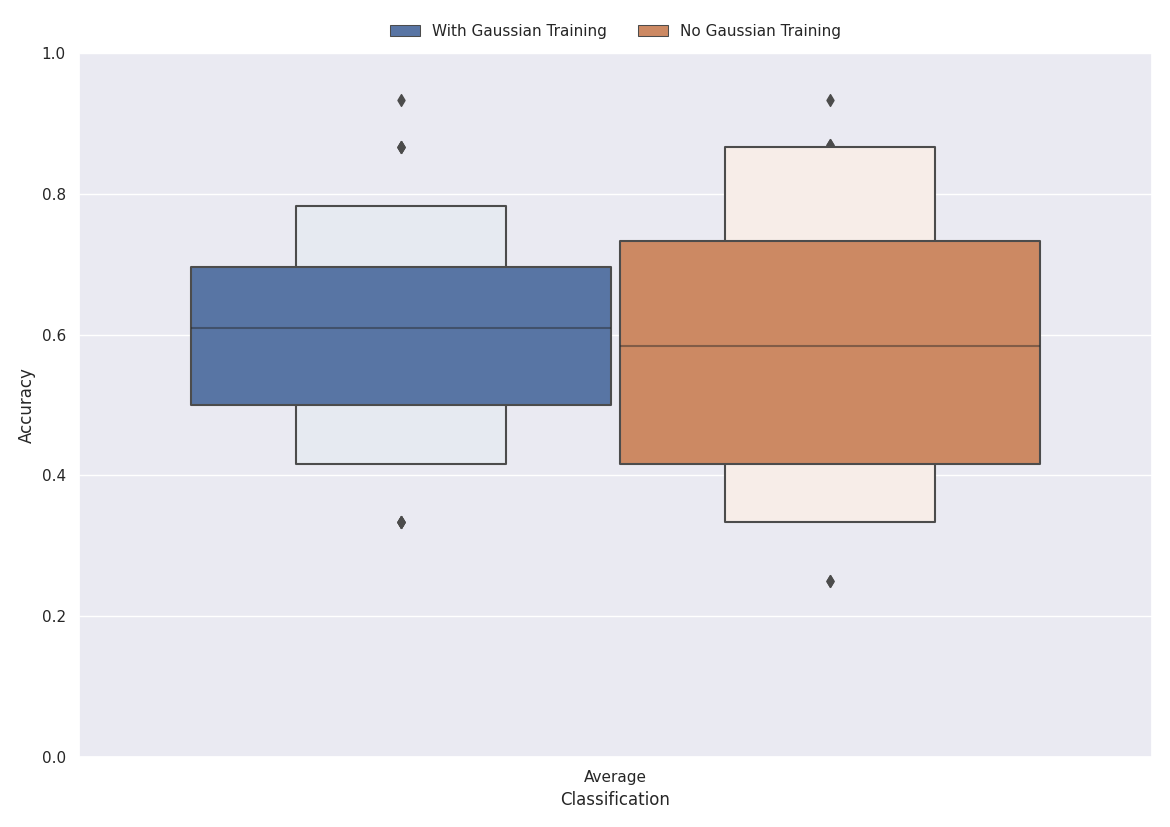}
    \caption{Performance with and without training of the Gaussian distributions on the ripening fruit data set.}
    \label{fig:abl_gaussian_training}
\end{figure}

\subsection{Training of the Gaussian Distributions}
We evaluated whether training the Gaussian distributions is necessary for the fruit ripening data set. As a comparison, we used the distribution previously mentioned, which should cover the whole inspected wavelength range in an evenly distributed way. The result is visible in Fig. \ref{fig:abl_gaussian_training}. There is a clear improvement in performance and stability.
\begin{figure}
    \centering
    \includegraphics[width=\columnwidth]{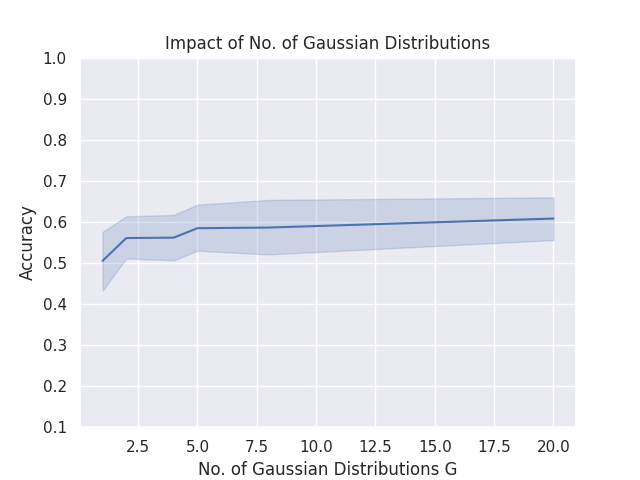}
    \centering
    \caption{Performance of for different number of WROIs on the ripening fruit data set with recordings of the Specim FX10.}
    \label{fig:deephs_abls_num}
\end{figure}

\subsection{Impact of Parameter $G$}
\label{sec:abl_g}
Fig. \ref{fig:deephs_abls_num} visualizes the impact of the parameter $G$, which defines the number of possible WROIs, on the performance of our model. This was tested on the ripening fruit data set and averaged over all setups (fruit type and classification type). A clear trend is visible. Increasing the number of Gaussian distributions increases the performance of the model. Two giant steps are visible: a Plateau between $G=2$ and $G=4$ and for $G\ge5$. The increase is continuous for $G>5$, but only slightly. Each jump in performance shows that the additional WROI provides helpful new information. Overall, the hyperparameter $G$ seems stable above the threshold of 5. So, for a ripeness prediction for different fruit types, $G=5$ is recommendable. For avocados, the classification seems more straightforward, and therefore already, $G=4$ are sufficient (this was shown in section 5.1. of the main work). $G=5$ also worked well for the second application B. As a result, $G=5$ seems a reliable first choice, which the first training results can optimize.

\begin{figure}
	\centering
	\includegraphics[width=\columnwidth]{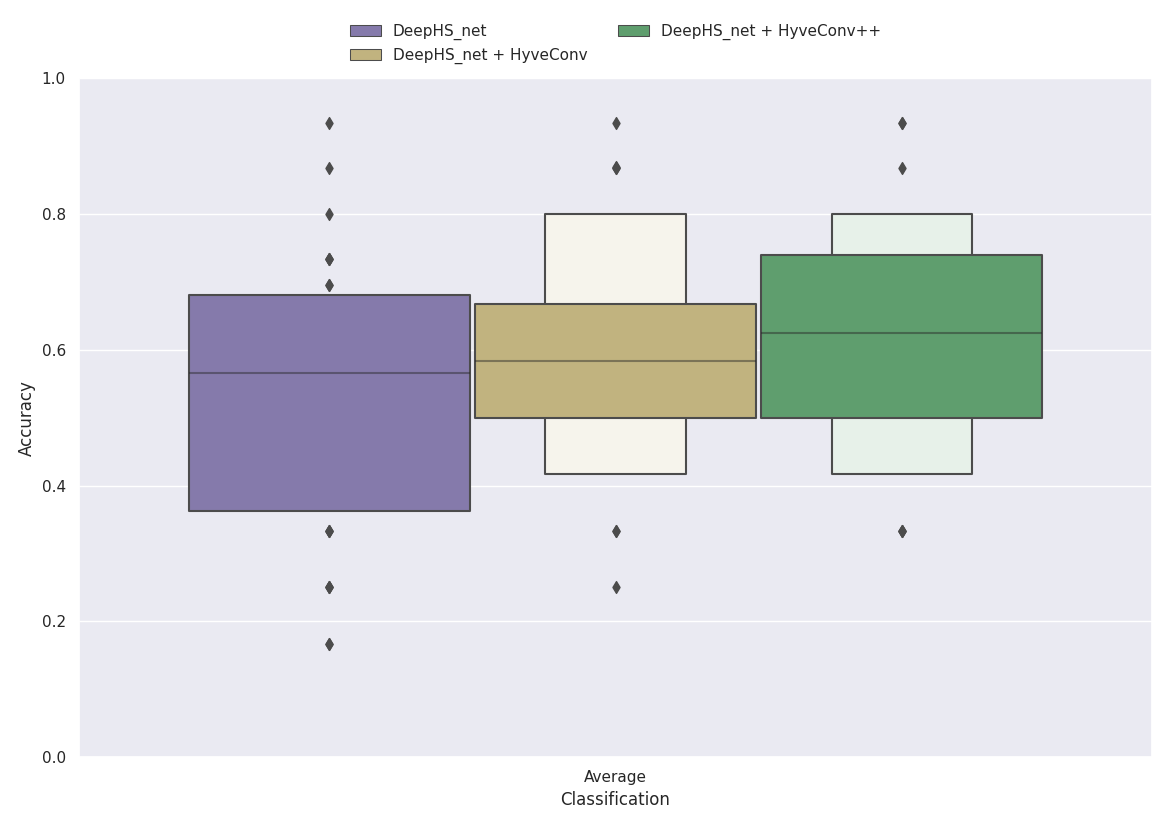}
	\centering
	\caption{Performance of different approaches on the ripening fruit data set with recordings of the Specim FX10.}
	\label{fig:deephs_abls_share}
\end{figure}

\subsection{Impact of the Method Extension}
We checked whether the proposed extension of the basic method is useful. We compare the baseline model (DeepHS\_net) with the basic method (HyveConv) and the extended method (HyveConv++). We did these experiments on the fruit ripening data set and averaged over all setups. The recordings of the Specim FX 10 were only used.
Fig. \ref{fig:deephs_abls_share} shows the performance of the different approaches. We tested the baseline model (DeepHS\_net) with and without our camera agnostic convolution layer. Further, the impact of the proposed extension is shown. The continuous definition of the input channel space of the convolution (HyveConv) boosts the accuracy by around 2\%. Allowing the convolution to share features through different output channels and the whole convolution layers(HyveConv++) boosts the model's performance by an additional 4\%. Therefore, the extension seems reasonable.

\section{Visualization of the HRSS prediction result}
\begin{figure*}[h]
	\begin{subfigure}{\textwidth}
		\centering
		\begin{subfigure}{0.39 \textwidth}
			\includegraphics[width=\columnwidth]{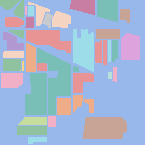}
			\caption{Prediction}
		\end{subfigure}
		\begin{subfigure}{0.39 \textwidth}
			\includegraphics[width=\columnwidth]{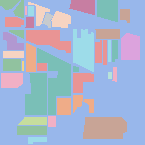}
			\caption{Ground Truth}
		\end{subfigure}
		\caption*{Prediction of DeepHS\_net + HyveConv++ and ground truth for the segmentations mask of the Indian pines data set.}
	\end{subfigure}
	\begin{subfigure}{\textwidth}
		\centering
		\begin{subfigure}{0.31 \textwidth}
			\includegraphics[width=\columnwidth]{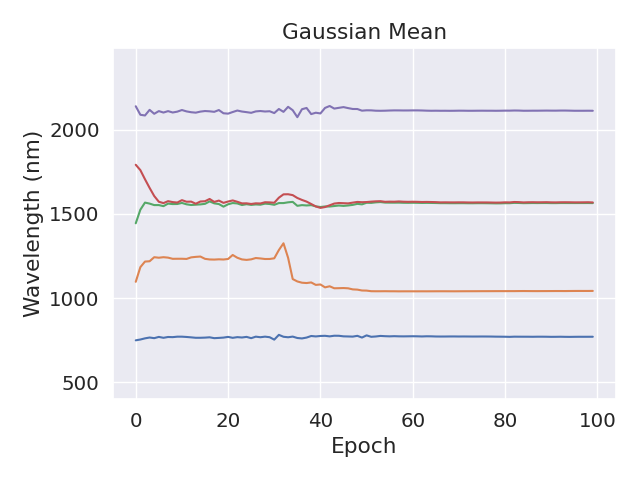}
			\caption{}
		\end{subfigure}
		\begin{subfigure}{0.31 \textwidth}
			\includegraphics[width=\columnwidth]{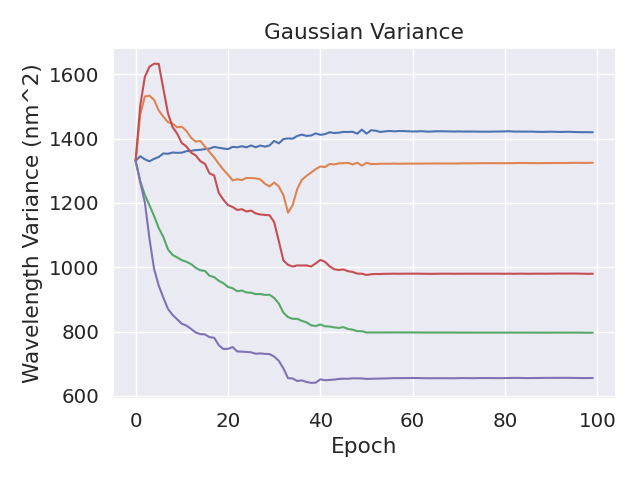}
			\caption{}
		\end{subfigure}
		\begin{subfigure}{0.31 \textwidth}
			\includegraphics[width=\columnwidth]{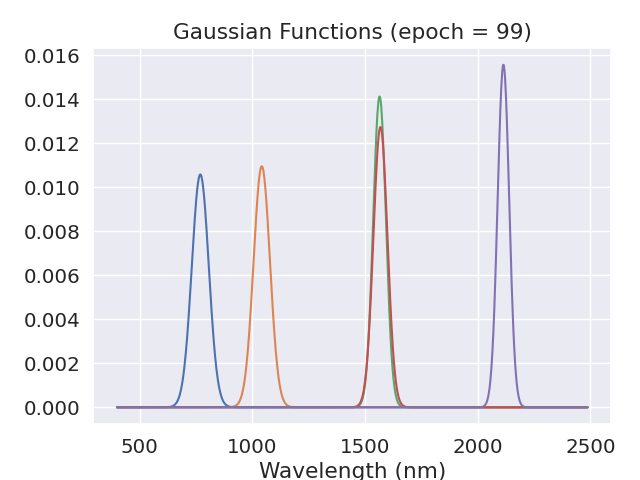}
			\caption{}
		\end{subfigure}
		\begin{subfigure}{\textwidth}
			\centering
			\includegraphics[width=0.9\textwidth]{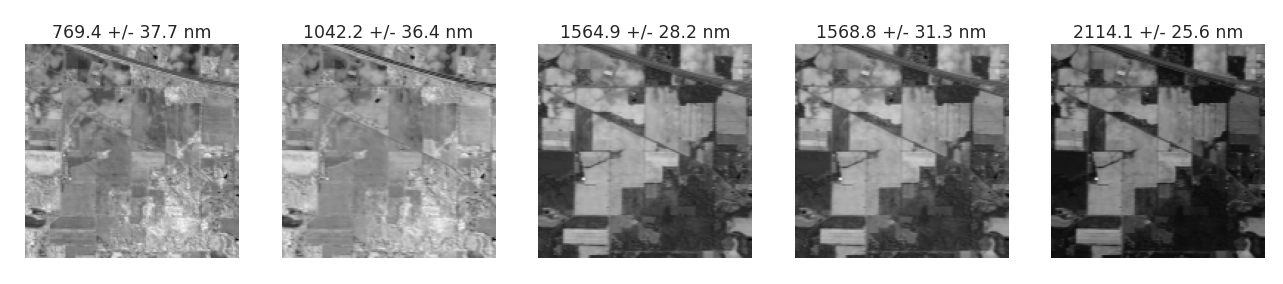}
			\caption{}
		\end{subfigure}
		\caption*{Training of the Gaussian distributions for the Indian pines data set. (a) and (b) show the development of the mean and variance over training epochs. (c) shows the final Gaussian distributions, and these are applied as filters in (d).}
	\end{subfigure}
	\caption{Indian pines}
	\label{fig:indian_pines_result}
\end{figure*}

\begin{figure*}[h]
	\begin{subfigure}{\textwidth}
 \centering
 \begin{subfigure}{0.20 \textwidth}
     \includegraphics[width=\columnwidth]{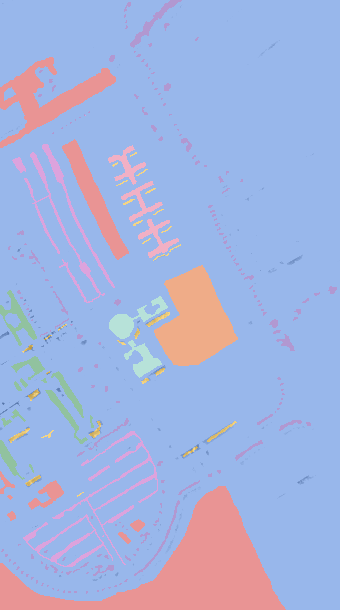}
     \caption{Prediction}
 \end{subfigure}
 \begin{subfigure}{0.20 \textwidth}
     \includegraphics[width=\columnwidth]{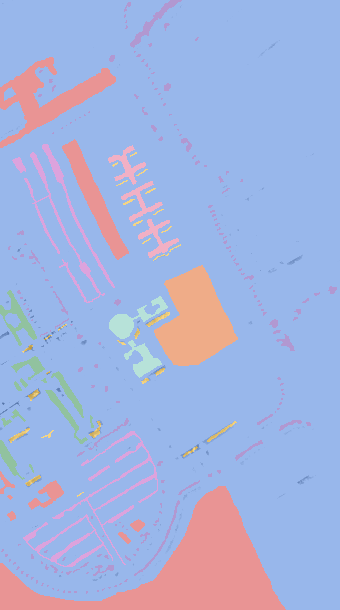}
     \caption{Ground Truth}
 \end{subfigure}
 \caption*{Prediction of DeepHS\_net + HyveConv++ and ground truth for the segmentations mask of the University of Pavia data set.}
\end{subfigure}
\begin{subfigure}{\textwidth}
 \centering
 \begin{subfigure}{0.31 \textwidth}
     \includegraphics[width=\columnwidth]{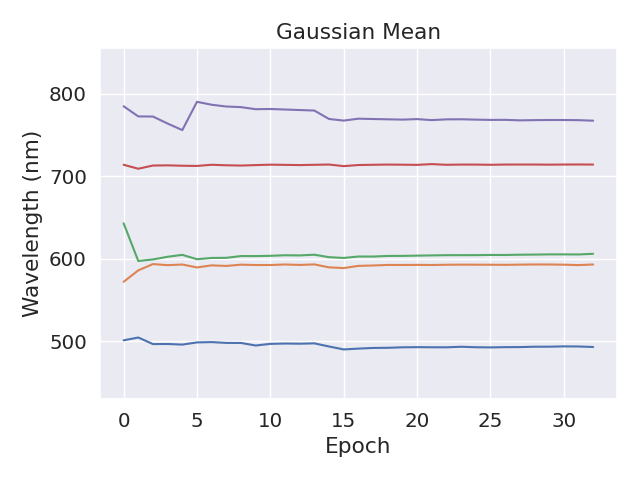}
     \caption{}
 \end{subfigure}
 \begin{subfigure}{0.31 \textwidth}
     \includegraphics[width=\columnwidth]{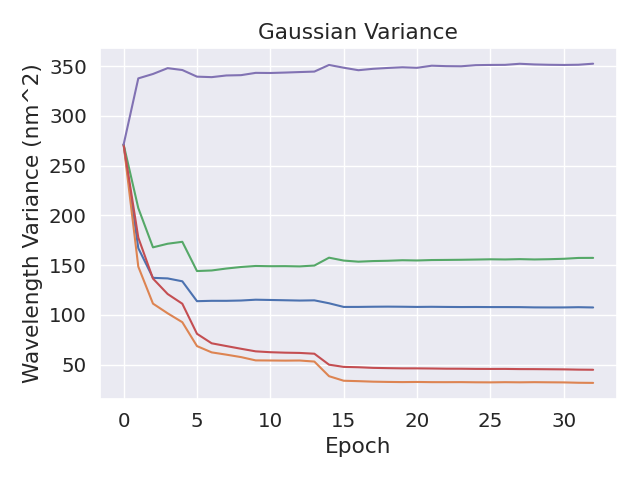}
     \caption{}
 \end{subfigure}
 \begin{subfigure}{0.31 \textwidth}
     \includegraphics[width=\columnwidth]{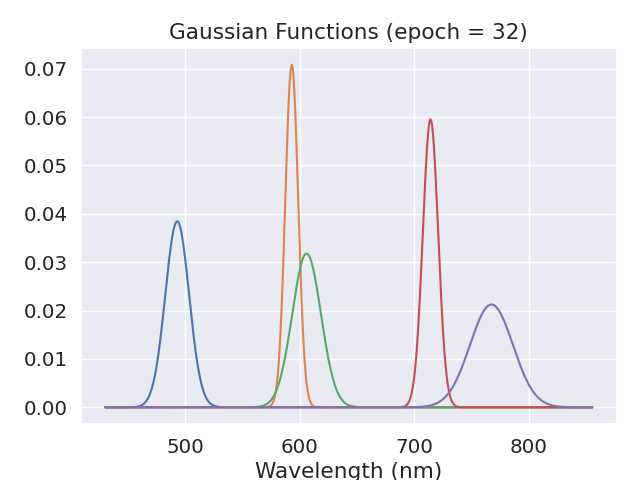}
     \caption{}
 \end{subfigure}
 \begin{subfigure}{\textwidth}
 \centering
 \includegraphics[width=0.8\textwidth]{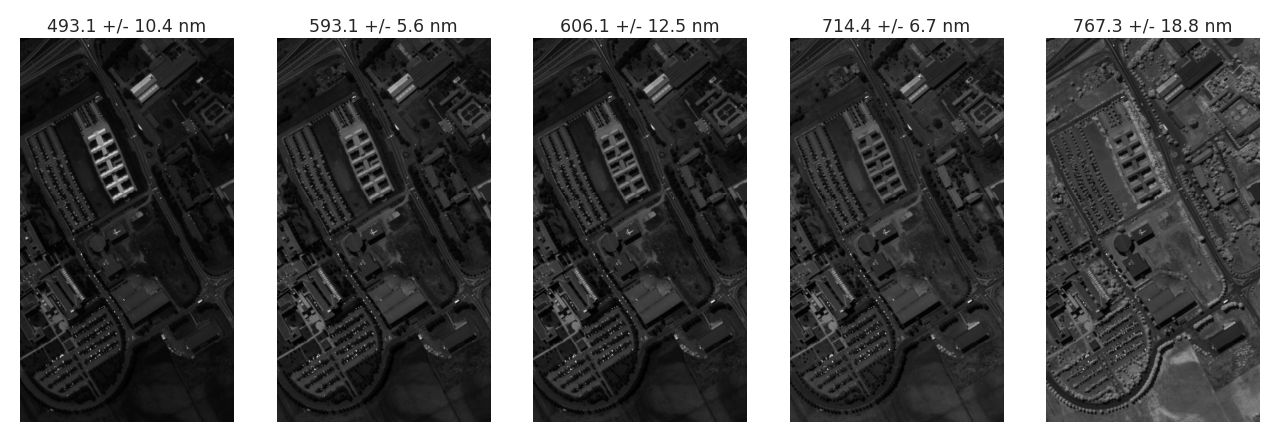}
 \caption{}
 \end{subfigure}
 \caption*{Training of the Gaussian distributions for the Pavia University data set. (a) and (b) show the development of the mean and variance over training epochs. (c) shows the final Gaussian distributions, and these are applied as filters in (d).}
	\end{subfigure}
  \caption{University of Pavia}
  \label{fig:paviau_result}
\end{figure*}

\begin{figure*}[h]
	\begin{subfigure}{\textwidth}
		\centering
		\begin{subfigure}{0.20 \textwidth}
			\centering
			\includegraphics[width=\columnwidth]{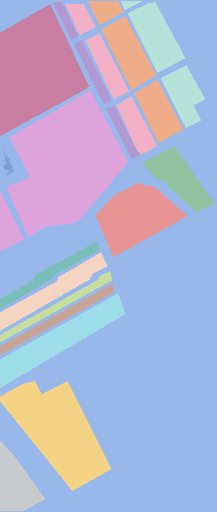}
			\caption{Prediction}
		\end{subfigure}
		\begin{subfigure}{0.20 \textwidth}
			\centering
			\includegraphics[width=\columnwidth]{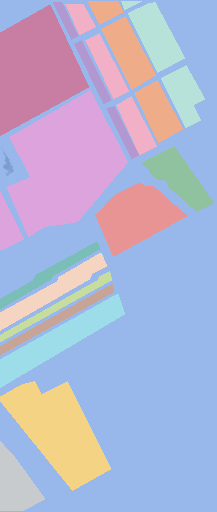}
			\caption{Ground Truth}
		\end{subfigure}
		\caption*{Prediction of DeepHS\_net + HyveConv++ and ground truth for the segmentations mask of the Salinas data set.}
	\end{subfigure}

	\begin{subfigure}{\textwidth}
		\centering
		\begin{subfigure}{0.31 \textwidth}
			\includegraphics[width=\columnwidth]{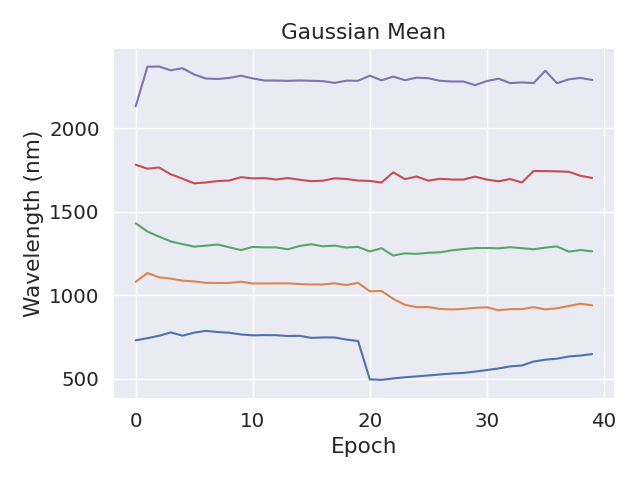}
			\caption{}
		\end{subfigure}
		\begin{subfigure}{0.31 \textwidth}
			\includegraphics[width=\columnwidth]{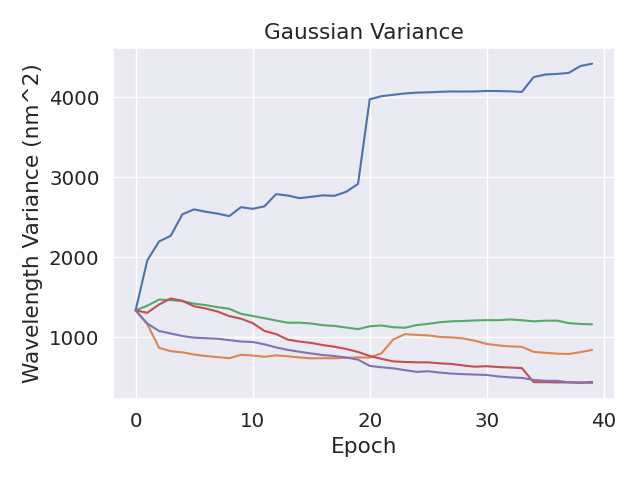}
			\caption{}
		\end{subfigure}
		\begin{subfigure}{0.31 \textwidth}
			\includegraphics[width=\columnwidth]{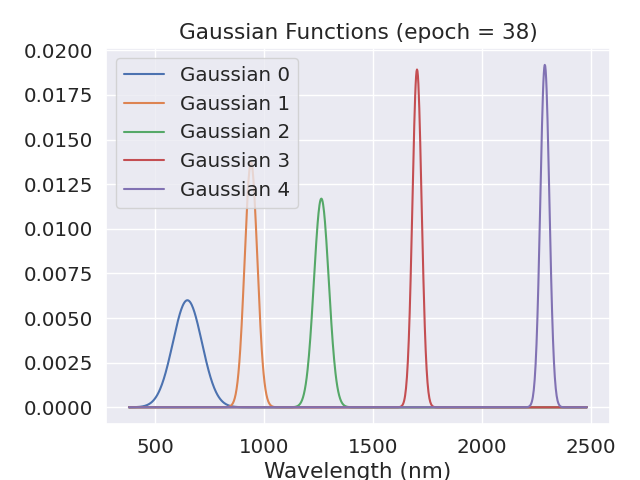}
			\caption{}
		\end{subfigure}
		\begin{subfigure}{\textwidth}
			\centering
			\includegraphics[width=0.65\textwidth]{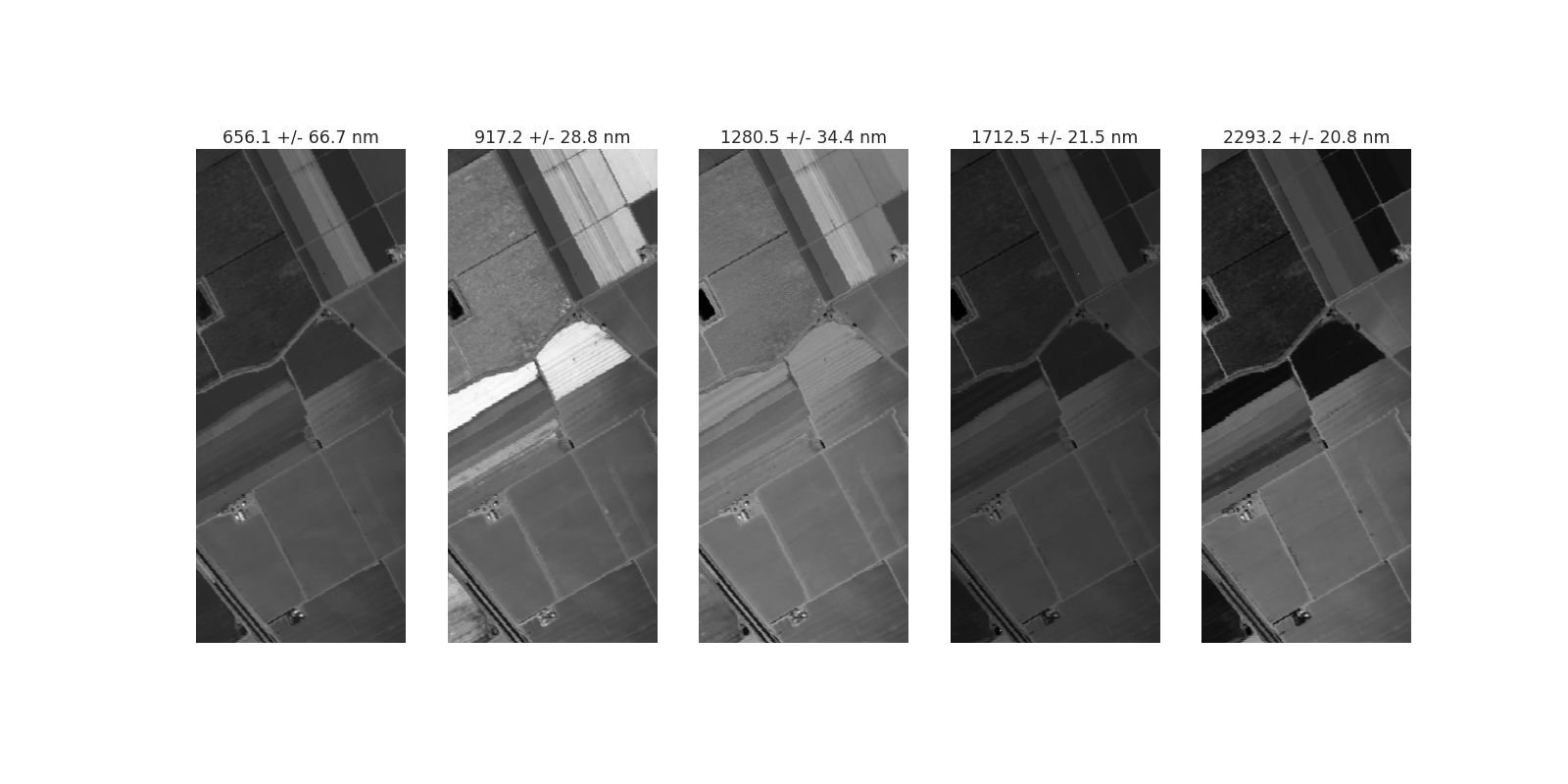}
			\caption{}
		\end{subfigure}
		\caption*{Training of the Gaussian distributions for the Salinas data set. (a) and (b) show the development of the mean and variance over training epochs. (c) shows the final Gaussian distributions, and these are applied as filters in (d).}
		
	\end{subfigure}
	\caption{Salinas}
	\label{fig:salinas_result}
\end{figure*}

For the HRSS data set, a visual comparison of the prediction and the ground truth is common, which can be found in Fig. \ref{fig:salinas_result} (Salinas), Fig. \ref{fig:paviau_result} (University of Pavia) and Fig. \ref{fig:indian_pines_result} (Indian pines). Especially for the Indian pines data set, some prediction errors of the model are visible. The corners of some areas are falsely classified.\\
Further, we provide for each model the training procedure of the camera filters and the final camera filters.

{\small
\bibliographystyle{ieee_fullname}
\bibliography{egbib}
}